\pdfoutput=1

\documentclass[11pt]{article}

\usepackage{acl}

\usepackage{times}
\usepackage{latexsym}

\usepackage[T1]{fontenc}
\usepackage{amssymb}

\usepackage[utf8]{inputenc}

\usepackage{microtype}

\usepackage{bbding} 
\usepackage{makecell}
\usepackage{multirow}
\usepackage{hyperref}
\usepackage{graphicx}
\usepackage{amsmath}

\usepackage{subfigure}
\usepackage{caption}
\usepackage{subcaption}
\usepackage{bm}
\usepackage{authblk}

\title{SpikeVoice: High-Quality Text-to-Speech Via Efficient Spiking Neural Network}

\author[1,2]{Kexin Wang}
\author[1,2]{Jiahong Zhang}
\author[1,2]{Yong Ren}
\author[1]{Man Yao}
\author[1,2]{Di Shang}
\author[1,2]{Bo Xu}
\author[1,2,3*]{Guoqi Li}
\affil[1]{Institute of Automation, Chinese Academy of Sciences, Beijing, China}
\affil[2]{School of Artificial Intelligence, University of Chinese Academy of Sciences, Beijing, China}
\affil[3]{Key Laboratory of Brain Cognition and Brain-inspired Intelligence Technology, Beijing, China}
\affil{wangkexin2021@ia.ac.cn}
\affil[*]{guoqi.li@ia.ac.cn (corresponding author)}

\begin{document}
\maketitle
\begin{abstract}
Brain-inspired Spiking Neural Network (SNN) has demonstrated its effectiveness and efficiency in vision, natural language, and speech understanding tasks, indicating their capacity to "see", "listen", and "read". In this paper, we design \textbf{SpikeVoice}, which performs high-quality Text-To-Speech (TTS) via SNN, to explore the potential of SNN to “speak”. A major obstacle to using SNN for such generative tasks lies in the demand for models to grasp long-term dependencies. The serial nature of spiking neurons, however, leads to the invisibility of information at future spiking time steps, limiting SNN models to capture sequence dependencies solely within the same time step.
We term this phenomenon "partial-time dependency". To address this issue, we introduce Spiking Temporal-Sequential Attention ($\textbf{STSA}$) in the SpikeVoice.
To the best of our knowledge, SpikeVoice is the first TTS work in the SNN field. We perform experiments using four well-established datasets that cover both Chinese and English languages, encompassing scenarios with both single-speaker and multi-speaker configurations. The results demonstrate that SpikeVoice can achieve results comparable to Artificial Neural Networks (ANN) with only $\bm{10.5\%}$ energy consumption of ANN. Both our demo and code are available as supplementary material.

\end{abstract}

\section{Introduction}

Since the advent of Artificial Neural Networks (ANN), remarkable achievements have been made in the field of image~\cite{clip,dert,swin,Speck}, natural language~\cite{transformer,bert,gpt}, and speech~\cite{wav2vec2,hubert}. In recent years, with the success of large language models~\cite{gpt4,palm2,llama2,blip2,emu,whisper}, there has been a notable upward trend in energy consumption. At the same time, Spiking Neural Network (SNN), inspired by the biological nervous system and recognized as the third generation of neural networks~\cite{spiking_first}, employs spiking neurons~\cite{HH,IF1999,psn} with charge-fire-reset temporal dynamic. The temporal dynamic makes SNN to exhibit the event-driven feature of sparse firing and the binary spike communication feature between neurons using 0s and 1s, providing a distinct advantage in energy efficiency~\cite{spiking_energy}.

Recently, SNN has achieved remarkable progress on several tasks, such as object detection and image classification~\cite{image_detect1,image_detect2,spike-driven,AttentionSNN,spikedrivenv2}, speech recognition~\cite{spike_speech1,spike_speech2}, and text classification tasks~\cite{spikebert,spike_text}. It is the success of these tasks that have led us to believe that SNN has preliminarily acquired the abilities of "seeing", "listening", and "reading". However, applying SNN to generative tasks encounters some obstacles, particularly in addressing the challenge of SNN capturing long-term dependencies. As mentioned above, spiking neurons have a temporal dynamic of charge-fire-reset. Such a serial process hinders the capture of information from future time steps in the spiking temporal dimension. Existing SNN models performing attention operations in the spiking sequential dimension can only establish sequence dependencies within the same time step or, in other words, among partial binary embedding~\cite{spikebert,spikeclip}, hindering the establishment of long-term dependencies. We term this phenomenon as "partial-time dependency". 

In this paper, we introduce SpikeVoice, a high-quality Text-To-Speech (TTS) model with a Transformer-based SNN framework~\cite{transformer} solving the "partial-time dependency" problem, and successfully explore the potential of SNN to "speak". To address the issue of "partial-time dependency", we propose Spiking Temporal-Sequential Attention (STSA) in SpikeVoice. STSA performs temporal-mixing in the spiking temporal dimension to capture information from future time steps, enabling access to the global information of binary embedding at each spiking time step. After time-mixing, STSA performs sequential-mixing in the spiking sequential dimension to integrate contextual information. Furthermore, we implement SpikeVoice in a spike-driven manner with the Leaky Integrate-and-Fire (LIF)~\cite{spiking_first} neurons, fully harnessing the energy efficiency of SNN. Spike-driven denotes the concurrent existence of both the binary spike communication feature and the event-driven feature. To the best of our knowledge, SpikeVoice is the first TTS model within the SNN framework, which not only promotes the development of SNN in generative tasks but also expands the scope of the SNN model in practical applications.

The main contributions are summarized as follows:
\begin{itemize}
    \item To the best of our knowledge, SpikeVoice is the first TTS model within the SNN framework that endows SNN with the "speaking" capability, enabling high-quality speech synthesis and filling the blank of speech synthesis in the SNN field. 
    \item In SpikeVoice, we introduce STSA, where the temporal-mixing in the spiking temporal dimension enables the access to the global information of binary embedding at each spiking time step, resolving the issue of "partial-time dependency" caused by the serial spiking neurons.
    \item The results reveal that SpikeVoice achieves synthesis performance close to ANN in both English and Chinese scenarios with both single-speaker and multi-speaker configurations. Remarkably, the energy consumption of SpikeVoice is merely $10.5\%$ of ANN, alleviating the high energy consumption issue associated with ANN.
\end{itemize}




\begin{figure*}[h]
  \centering
  \includegraphics[width=0.95\linewidth]{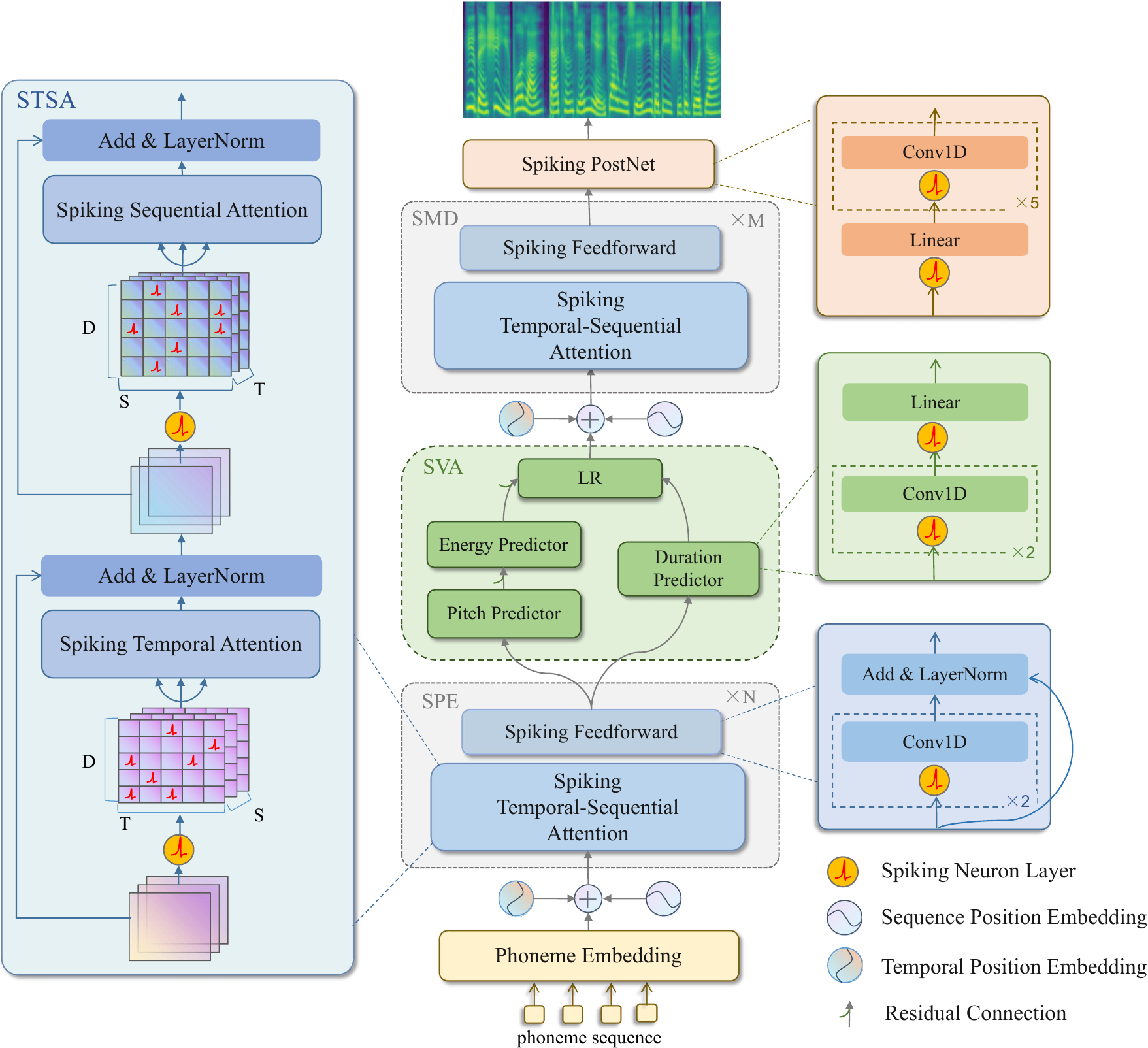}
  \caption{The overview model structure of SpikeVoice. In the figure, the left part represents the Spiking Temporal-Sequential Attention (STSA). In the middle part, from bottom to top, are the Spiking Phoneme Encoder (SPE), Spiking Variance Adapter (SVA), and Spiking Mel Decoder (SMD) with the topmost part represents the output Mel-Spectrogram. On the right part, the green module represents the predictor within the Spiking Variance Adapter, the blue module represents Spiking FeedForward, and the orange module indicating Spiking PostNet. }
  \label{spikevoice}
  \end{figure*}

\section{Related work}
\textbf{Transformers in SNN:} Training in SNN is primarily categorized into two methods: ANN-to-SNN conversion (ANN2SNN)~\cite{optimal,deng2021optimal,han2020rmp} and surrogate training~\cite{wu2018spatio,shrestha2018slayer,stbp,duan2022temporal}. Leveraging ANN2SNN, \cite{convert} integrates the Transformer architecture into SNN. Nevertheless, this approach demands dozens or even hundreds of time steps to attain satisfactory performance. Spikeformer~\cite{spikeformer} conducts direct training of the Transformer within the SNN framework and achieves state-of-the-art performance on ImageNet with just four time steps. However, it doesn't fully harness the energy-efficient advantages of SNN due to the presence of  Multiply-and-Accumulate (MAC) operations. Spike-driven Transformer~\cite{spike-driven} incorporates the spike-driven paradigm into Transformer architecture and introduces the Spike-Driven Self-Attenton (SDSA)~\cite{spike-driven}. SDSA utilizes sparse additive operations as a replacement for multiplication operations in attention mechanisms, effectively addressing the issues present in Spikeformer related to MAC operations. SpikeGPT~\cite{spikegpt} is the first to introduce text generation tasks into the SNN framework. However, it still does not make full of the energy-efficient capabilities of SNN. 

\textbf{Transformers in TTS:} Tactron2~\cite{tacotron2} employs RNN~\cite{rnn} for speech synthesis which results in low training efficiency and struggles to establish long-term dependencies. To address these issues, Transformer-TTS~\cite{transformer-tts} introduces an autoregressive TTS model that combines Tactron2 with the Transformer, enhancing training efficiency while capturing long-term dependencies. However, autoregressive TTS models often suffer from slow synthesis speed and less robust speech synthesis. FastSpeech~\cite{fastspeech}, on the other hand, utilizes knowledge distillation during training to build a non-autoregressive TTS model, yet the training process can be complicated. FastSpeech2~\cite{fastspeech2} simplifies the training process by removing knowledge distillation from the FastSpeech training pipeline and adopting the end-to-end training approach, effectively addressing the issue of the extended training duration associated with FastSpeech.

\begin{table*}
\setlength{\tabcolsep}{0.3cm}
\centering
\begin{tabular}{c|ccc}
\hline
& & SpikeVoice & FastSpeech2\\
\hline
\multirow{5}{*}{\makecell[c]{STSA/Attention}} &$Q,K,V$ &  $T\Bar{R}_{t/s}*E_{add}* 3ND^2$ & $E_{mac}* 3ND^2$\\
&$F(Q,K,V)$ & $T\hat{R}_{t/s}*E_{add}* ND$ & $E_{mac}* ND^2$\\
&$Linear_0$ & $TR_{mlp_1}*E_{add}*FLP_{mlp_0}$ & $E_{mac}*FLP_{mlp_0}$\\
& $Scale$ & - & $E_m*N^2$ \\
& $Softmax$ & - & $E_{mac}*2N^2$ \\
\hline
\multirow{1}{*}{\makecell[c]{Spiking Feedforward}}  & $Conv\_Layer_{0/1}$ & $TR_{c_0/c_1}*E_{add}*FLP_{c_0/c_1}$ & $E_{mac}*FLP_{c_0/c_1}$ \\
\hline
\multirow{2}{*}{\makecell[c]{Predictors}} & $Conv\_Layer_{2/3}$ & $TR_{c_2/c_3}*E_{add}*FLP_{c_2/c_3}$ & $E_{mac}*FLP_{c_2/c_3}$ \\
& $Linear_1$ & $TR_{mlp_1}*E_{add}*FLP_{mlp_1}$ & $E_{mac}*FLP_{mlp_1}$\\
\hline
\multirow{2}{*}{\makecell[c]{Spiking PostNet}} & $Linear_2$ & $TR_{mlp_2}*E_{add}*FLP_{mlp_2}$ & $E_{mac}*FLP_{mlp_2}$\\
& $Conv\_Layer_{4-9}$ & $TR_{c_4-c_9}*E_{add}*FLP_{c_4-c_9}$ & $E_{mac}*FLP_{c_4-c_9}$ \\
\hline
\end{tabular}
\caption{The energy consumption estimation of the main components. $T$ is the total time steps, and $R$ denotes the firing rates of spike tensors. $E_{add}=0.9pJ$ and $E_{mac}=4.6pJ$ are the energy consumption of add and MAC operations at 45nm process nodes for full precision~(FP32) SynOps. $N$ is the length of sequences, and $D$ represents the number of channels. $FLP_c$ and $FLP_{mlp}$ are FLOPs of Conv layers and MLP layers.}
\label{energy consumption}
\end{table*}

\section{Method}

In this study, we propose SpikeVoice, the first spike-driven TTS model. The overall model architecture is illustrated in Fig.\ref{spikevoice}. The Spiking Phoneme Encoder (SPE) performs binary embedding on the input phoneme embedding sequence and generates high-level spiking phoneme representations. The Spiking Variance Adaptor (SVA) enhances the spiking phoneme representations by incorporating variance information related to duration, pitch, and energy.  Finally, the Spiking Mel Decoder (SMD) and Spiking PostNet generate Mel-Spectrograms in a non-autoregressive manner. In the following sections, we will first introduce the LIF neurons, and then introduce the components of SpikeVoice.

\subsection{Leaky Integrate-and-Fire Neuron}
\begin{figure}[h]
  \centering
  \includegraphics[width=0.85\linewidth]{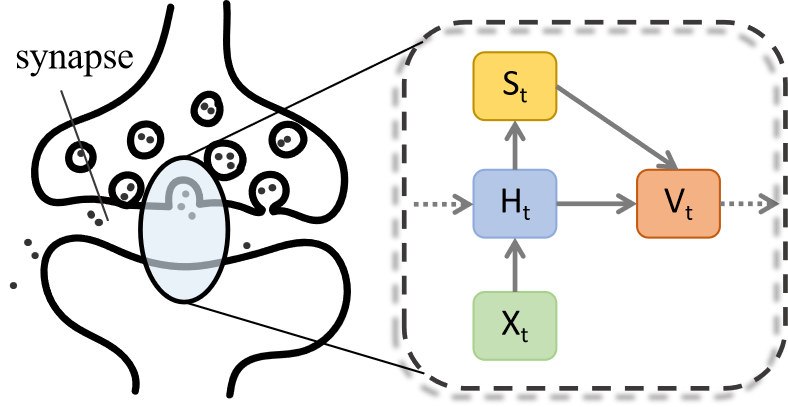}
  \caption{The LIF neuron layer.}
  \label{lif}
  \end{figure}

The LIF neuron is a biologically inspired spiking neuron having the charge-fire-reset biological neuronal dynamics as shown in Fig.\ref{lif}. The working process of LIF neuron can be described as:
\begin{align}
\label{neuron_lif}
    &H_t= V_{t-1}+\frac{1}{\tau}(X_t-(V_{t-1}-V^{re}))\\
    \label{neuron_lif2}
    &S_t=\Theta(H_t-V^{th})\\
    \label{neuron_lif3}
    &V_t=V^{re}S_t+H_t(1-S_t)
\end{align}

Eq.(\ref{neuron_lif}) to (\ref{neuron_lif3}) respectively represent the charging, firing, and membrane potential resetting of LIF. $X_t$ denotes the input current at time $t$, $H_t$ signifies the membrane potential after charging, $S_t$ represents the spike tensor at time $t$, $\Theta$ represents the step function, $V^{th}$ denotes the firing threshold, $V^{re}$ is the reset membrane potential, and $V_t$ signifies the membrane potential after resetting. 

\subsection{SpikeVoice}
\textbf{Temporal-Sequential Embedding:} 
At spiking temporal wise, we first expand the phoneme embedding sequence $z$ to $T$ time steps. In order to incorporate the position information with STSA, we then apply position embedding in both the spiking temporal dimension and the phoneme sequential dimension. 
\begin{align}
&x^0_{(t,l)}=z_{(t,l)}+e^{tem}_{(t,)}+e^{seq}_{(,l)}
\end{align}
where $x^0\in\mathcal{R}^{T\times L\times D}$ will be taken as the input to Spiking Phoneme Encoder. $L$ represents the length of the phoneme sequence, $D$ denotes the size of embedding dimension, $t\in\{1,\ldots,T\}$ and $l\in\{1,\ldots,L\}$. $e^{tem}_{(t,)}$ and $e^{seq}_{(,l)}$ are the position embedding of time step $t$ at temporal wise and position $l$ at sequence wise. 

\textbf{Spiking Phoneme Encoder:} Spiking Phoneme Encoders are composed of a stack of $N$ identical layers, each of which consists of an STSA module and a Spiking FeedForward module. As shown on the right side of Fig.\ref{spikevoice}, each Spiking FeedForward module consists of two stacked 1D-Convolution layers. To ensure the energy efficiency of SpikeVoice, we introduce a LIF neuron layer before each 1D-Convolution layer, to convert continuous inputs into sparse spiking tensors. Then the high-level spiking phoneme representations $x^n$ of layer $n$ can be obtained as:
\begin{align}
u^n&=STSA(x^{n-1})\\
x^n&=LN(u^n+f(u^n))\\
f(\cdot)&=[Conv(\mathcal{SN(\cdot)})]_2
\end{align}
where $\mathcal{LN}$ is layer nomalization, $\mathcal{SN}$ refers to the LIF neuron layer depicted in Eq.(\ref{neuron_lif})-(\ref{neuron_lif3}). $f(\cdot)$ represents the stacked 1D-Convolution and LIF neuron layers, $u^n$ is the membrane potential output  of STSA. 

\textbf{Spiking Temporal-Sequential Attention:} 
As illustrated in the left block of Fig.\ref{spikevoice}, STSA is composed of a Spiking Temporal Attention and a Spiking Sequential Attention. Due to the serial nature of LIF neurons, it results in the inability to capture information from future time steps along the spiking temporal dimension and leads to the issue of "partial-time dependency". Therefore, we propose the Spiking Temporal Attention to perform temporal-mixing over the spiking temporal dimension obtaining the global information of binary embedding.

Taking STSA in layer $n$ of Spiking Phoneme Encoder as an example, initially, we perform binary embedding on the output of layer $n-1$ to obtain the sparse spiking hidden representation $s^{n}=\mathcal{SN}(x^{n-1})$, $s^{n} \in \mathcal{R}^{T\times L\times D}$. Along the spiking temporal dimension $T$ the binary embedding of each token can be obtained. The Spiking Temporal Attention can be depicted as:
\begin{align}
\mu^{n}&=\mathcal{SN}({BN}(W^{n,tem}_{\mu}{s}^{n}))\\
\label{att}
s^{n}_{(t,:)}&=\mathcal{SN}(\Sigma_c(q^{n}_{(t,:)}\odot k^{n}_{(t,:)}))\odot v^{n}_{(t,:)}\\
\sigma^{n}&={LN}(x^{n-1}+Linear({s}^{n}))
\end{align}
 where $\mu\in\{q,k,v\}$, $\mathcal{BN}$ represents Batch Normalization, and $W^{n,tem}_\mu$ is a learnable matrix for Spiking Temporal Attention. For vanilla attention can introduce MAC operations to the SpikeVoice, we utilize the SDSA~\cite{spike-driven} in Eq.(\ref{att}) as a substitute for vanilla attention. $\odot$ is the Hadamard product and $\Sigma_c$ means sum up in column-wise.  $s^{n}_{(t,:)}$ denotes the spiking tensor at time step $t$, which is the output of attention computing on spiking temporal wise. $\sigma^{n}$ represents the membrane potential output of Spiking Temporal Attention.

Then $s^n=\mathcal{SN}(\sigma^{n})$ will serve as the sparse input to Spiking Sequential Attention:
\begin{align}
\mu^{n}&=\mathcal{SN}({BN}(W^{n,seq}_{\mu}s^n))\\
s^{n}_{(:,l)}&=\mathcal{SN}(\Sigma_c(q^{n}_{(:,l)}\odot k^{n}_{(:,l)}))\odot v^{n}_{(:,l)}\\
u^{n}&={LN}(u^{n}+Linear(s^{n}))
\end{align}
where $s^{n}_{(:,l)}$ is the spiking tensor at position $l$ in the sequence wise. The computation process above can be easily extended to Spiking Mel Decoder.

\textbf{Spiking Variance Adaptor:} The Spiking Variance Adaptor takes the high-level spiking phoneme representations $x^{N}$ as its input. And then the Duration Predictor $P_d$, Energy Predictor $P_e$, and Pitch Predictor $P_p$ impart variance information to $x^{N}$. The predictors in Spiking Variance Adaptor all take an identical structure, shown in the green block on the right side of Fig.\ref{spikevoice}. Besides, We employ a residual connection around the Energy Predictor and Pitch Predictor. Finally, the Length Regulator $LR$ aligns the hidden sequence to the length of the Mel-Spectrogram:
\begin{align}
d &= P_{d}(x^{N})\\
u &= P_{e}(P_p(x^{N}))\\
{\{y^0_{(t,l^\prime)}\}}_{l^\prime=1,\ldots,L^\prime} &= LR\left(u_{(t,l)},d_{(l,)}\right)_{l=1,\ldots,L}
\end{align}
where $d \in R^L$  comprises the length of mel frames corresponding to each phoneme. $u$ represents the membrane potential incorporated the pitch and energy variance information.  
$\{y^0_{(t,l^\prime)}\}$ signifies the mel representations corresponding to ${u_{(t,l)}}$ after being extended by $d_{(l,)}$ times. $L^\prime$ represents the total length of the target Mel-Spectrogram.

\textbf{Spiking Mel Decoder and PostNet:} Spiking Phoneme Encoders are composed of a stack of $M$ identical layers, each of which also comprises an STSA and a Spiking FeedForward. The Spiking PostNet is designed to enhance the fine details of Mel-Spectrograms. LIF neuron layers are also added before each linear layer and 1D-convolution layer in the Spiking PostNet to ensure sparse inputs. 
Then the Mel-Spectrogram can be obtained as:
\begin{align}
y^{m}& =SFF(STSA(y^{m-1}))\\
O&=PostNet(y^{M})\\
O^{c}_{(l^\prime,)}&=\bar{y}^M_{(:,l^\prime)},\quad O^{f}_{(l^\prime,)}=\bar{O}_{(:,l^\prime)}
\end{align}
where $y^m$ is the output of the $m$th layer of Spiking Mel Decoder. To calculate the supervised loss with ground truth, we average the output at spiking temporal dimension as the predicted Mel-Spectrograms, and $\bar{\cdot}$ represents the average operation. We denote the Mel-Spectrograms obtained before the Spiking PostNet as $O^{c}$ and the output obtained from the Spiking PostNet as $O^{f}$. 

The loss function encompasses supervised losses using Mean Squared Error (MSE) for pitch, energy, and duration, as well as Mean Absolute Error (MAE) losses for both the coarse Mel-Spectrograms $O^{c}$ and the fine Mel-Spectrograms $O^{f}$.

\begin{table*}
\setlength{\tabcolsep}{0.001cm}
\centering
\begin{tabular}{c|ccc|ccc}
\hline
\multicolumn{7}{c}{Single-Speaker}\\
\hline
&\multicolumn{3}{c|}{LJSpeech}&\multicolumn{3}{c}{Baker}\\
\hline
Methods &WER$\downarrow$ & NISQA-V2$\uparrow$ & MOS$\uparrow$ & CER$\downarrow$ & NISQA-V2$\uparrow$ & MOS$\uparrow$\\
\hline
\textit{GT} &$6.39$ &$4.42$ &$4.75\pm.037$ &$12.25$ &$4.06$ & $4.30\pm.052$\\
\hline
\textit{FastSpeech2~\cite{fastspeech2}}&$7.98$ &\underline{4.13} &\underline{$4.10\pm.057$} &$13.18$ &{$3.80$} & $3.82\pm.089$\\
\hline
\textit{SpikeVoice-ATTN~\cite{spikeformer}}  &$8.39$ &$4.08$ &$3.69\pm.053$ &$13.16$ &$3.78$ & $3.52\pm.093$\\
\textit{SpikeVoice-SDSA~\cite{spike-driven}} &$8.70$ &$4.10$ &$3.63\pm.059$ &$12.96$ &$3.79$ &$3.46\pm.088$ \\
\textit{SpikeVoice-STSA~(ours)} &\bm{$7.93$} &\bm{$4.11$} &\bm{$4.06\pm.052$}  &\bm{$12.89$} &\bm{$3.80$} &\bm{$3.86\pm.076$}\\
\hline
\end{tabular}
\caption{Results on LJSpeech and Baker for experiments for single-speaker. \textit{GT} stands for ground truth, FastSpeech2 is the work of \cite{fastspeech2}. 
WER/CER and NISQA-V2 are the objective metric and MOS is the subjective metric. The best results of the SNN-based models are highlighted with \textbf{bold font}, and the \underline{underlined font} indicates that the performance of the ANN-based model is superior to the optimal performance of the SNN-based model.}
\label{single speaker}
\end{table*}

\section{Experiments}
We conducted experiments with SpikeVoice on single-speaker and multi-speaker datasets, encompassing both English and Chinese. The single-speaker datasets include LJSpeech~\cite{ljspeech} and Baker\footnote{https://www.data-baker.com/data/index/TNtts/}, while the multi-speaker datasets comprise LibriTTS~\cite{libritts} and AISHELL3~\cite{AISHELL-3}. In the following subsections, we present results on subjective and objective metrics for ground truth denoted as \textit{'GT'}, ANN baseline denoted as \textit{'FastSpeech2'}, SpikeVoice signified as \textit{'SpikeVoice-STSA'}, and SNN baselines: SpikeVoice with attention in Spikeformer replacing the STSA, which is denoted as \textit{'SpikeVoice-ATTN'} and SpikeVoice with only Spiking Sequential Attention, which denoted as \textit{'SpikeVoice-SDSA'}. Additionally, In Section \ref{visualize}, we perform visual analysis, and in Section \ref{balance}, we discuss the balance between SpikeVoice's energy consumption and the quality of synthesized speech.

\subsection{Datasets} 
For each of the datasets, we have randomly split the dataset into three sets: the training set, the validation, and the testing sets, both comprising 256 samples.

\textbf{LJSpeech} is a female single-speaker English monolingual dataset. It comprises a collection of 13100 utterances, each lasting between 1 to 10 seconds, amounting to roughly 24 hours of speech material. 

\textbf{Baker} is a female single-speaker Chinese dataset.  It encompasses a wide range of content domains, including news, novels, technology, and so on. In total, Baker comprises 10000 speech recordings, with approximately a total of 12 hours of speech material. 

\textbf{LibriTTS} comprises approximately 191 hours of speech with 1,160 speakers. We utilized the \textit{train-clean-360} set from LibriTTS. Within this subset, there are 430 female speakers and 474 male speakers. 

\textbf{AISHELL3} is a multi-speaker Chinese dataset, containing a total of approximately 85 hours of speech, recorded by 218 speakers.

\subsection{Experiments settings}
\textbf{Training Settings} SpikeVoice is stacked by $N=4$ Spiking Phoneme Encoders, a Spiking Variance Adaptor, and $M=6$ Spiking Mel Decoders. We transformed the raw speech in all the datasets into mel-spectrograms with a frame length of 1024 and a hop length of 256. The synthesized mel-spectrograms were uniformly converted into speech using the vocoder HiFiGAN~\cite{hifigan}. We performed the training on four Tesla V100-SXM2-32G GPUs with batch size 48. The optimization settings were in line with those defined in \cite{fastspeech2}. The implementation of the SNN framework in SpikeVoice is based on SpikingJelly~\cite{fang2023spikingjelly}.

\textbf{Evaluation Settings} We employed Word Error Rate (WER) for English and Character Error Rate (CER) for Chinese, along with NISQA-V2~\cite{nisqa}, as objective metrics to evaluate the quality of single-speaker speech synthesis. For multi-speaker synthesis, we additionally utilized Speaker Embedding Cosine Similarity (SECS) to gauge the similarity between the synthesized speech and the target speech in terms of the speaker's voice. Specifically, for WER, we utilized Hubert~\cite{hubert} for English ASR transcription and Wav2Vec2~\cite{wav2vec2} for Chinese ASR transcription. As for SECS, we employed the speaker encoder from the Resemblyzer\footnote{https://github.com/resemble-ai/Resemblyzer} toolkit to extract speaker embeddings and calculate cosine similarity. In assessing both single and multi-speaker synthesis, we relied on 5-scale Mean Opinion Scores (MOS) with $95\%$ confidence intervals as our subjective metric. To obtain these scores, we randomly selected 80 samples from each test set, and a total of 12 participants were asked to provide ratings for the synthesized speech.

\begin{table*}
\setlength{\tabcolsep}{0.1cm}
\centering
\begin{tabular}{c|cccc|cccc}
\hline
\multicolumn{9}{c}{Multi-Speaker}\\
\hline
&\multicolumn{4}{c|}{AISHELL3}&\multicolumn{4}{c}{LibriTTS}\\
\hline
Methods & WER$\downarrow$ & \makecell[c]{NISQA\\-V2}$\uparrow$ & SECS$\uparrow$ & MOS$\uparrow$ & CER$\downarrow$ & \makecell[c]{NISQA\\-V2}$\uparrow$ & SECS$\uparrow$ & MOS$\uparrow$\\
\hline
\textit{GT}&$5.36$ &$3.37$ &- &$4.48\pm .057$ &$5.07$ &$4.14$ &- &$4.46\pm .047$\\
\hline
\textit{FastSpeech2} &$6.36$ &$3.09$ &$0.849$ &\underline{$3.92\pm .059$}&$\underline{5.72}$ &$\underline{3.47}$ &$\underline{0.822}$&\underline{$3.43\pm .074$}\\
\hline
\textit{SpikeVoice-ATTN} &$7.13$ &$3.12$ &$0.841$& $3.55\pm .061$&$6.63$ &$3.42$ &$0.794$&$2.72\pm.089$ \\
\textit{SpikeVoice-SDSA} &$7.42$ &$3.12$ &$0.849$& $3.63\pm .058$ &$6.45$ &$3.40$ &$0.794$&$2.88\pm.066$ \\
\textit{SpikeVoice-STSA} &\bm{$6.32$} & \bm{$3.13$} &\bm{$0.850$}&\bm{$3.79\pm .056$}&\bm{$6.06$} & \bm{$3.43$} &\bm{$0.795$}&$\bm{3.32\pm.052}$ \\
\hline
\end{tabular}
\caption{Results on AISHELL3 and LibriTTS for experiments of multi-speaker. CER, NISQA-V2, and SCER are the objective metric and MOS is the subjective metric. The best results of the SNN-based models are highlighted with \textbf{bold font}, and the \underline{underlined font} indicates that the performance of the ANN-based model is superior to the optimal performance of the SNN-based model.}
\label{aishell3}
\end{table*}

\subsection{Performance on Single-Speaker}
As shown in Tab.\ref{single speaker}, we conducted experiments on the LJSpeech and Baker datasets, reflecting the synthesis quality of English and Chinese single-speaker respectively. 
 

For the objective metrics, SpikeVoice surpasses all the SNN and ANN baselines on the WER/CER metric and is the best-performing SNN-based model on NISQA.
These results demonstrate that the global information of temporal spike sequence in STSA contributes to the synthesis of higher-quality and clearer speech. 

For the subjective evaluation, SpikeVoice outperforms both \textit{SpikeVoice-ATTN} and \textit{SpikeVoice-SDSA}. 
 The difference in MOS scores between SpikeVoice and ANN is merely $0.04$ on LJSpeech and SpikeVoice even surpasses the ANN-based model on the Baker dataset, indicating that SpikeVoice's synthesis quality closely approaches that of ANN in terms of human perception. The results compared to \textit{SpikeVoice-SDSA} also confirm the effectiveness of temporal-mixing. 
 

\subsection{Model Performance on Multi-Speaker}
\label{multi}
In Tab.\ref{aishell3}, we respectively present the performance on the AISHELL3 and LibriTTS. In the multi-speaker experiments, we have additionally incorporated the SCER metric to assess the speaker similarity between synthesized speech and target speech. 


Compared to single-speaker, multi-speaker datasets present more challenges for SNN-based models. SpikeVoice with STSA remains the best-performing SNN-based model, however, the sparse nature of the spike tensor contributes to energy efficiency at the expense of information loss, leading to a performance gap of MOS scores between the SNN-based models and ANN-based models in multi-speaker datasets, which encompass richer information. Investigating strategies to minimize information loss in the context of spike tensors with low firing rates is worthwhile for future research.




\begin{table*}
\setlength{\tabcolsep}{0.15cm}
\centering
\begin{tabular}{c|ccccc|c}
\hline

Methods & \makecell[c]{Spike-Driven} & Complexity & Param &Time Step& E(pJ)&MOS\\
\hline
\textit{FastSpeech2} & \XSolidBrush & $O(N^2D)$ & $35.4$ & 1&2.14e11&\bm{$4.10\pm.057$} \\
\hline
\textit{SpikeVoice-ATTN} & \Checkmark & $O(TN^2D)$ & $35.4$ &4& 2.55e10&$3.69\pm.053$ \\
\textit{SpikeVoice-SDSA} & \Checkmark & $O(TND)$ & $35.4$ &4& 2.06e10&$3.63\pm.059$\\
\textit{SpikeVoice-STSA} & \Checkmark & $O(2TND)$ & $37.8$ &1& \textbf{8.84e09}& $3.61\pm.053$\\
\textit{SpikeVoice-STSA} & \Checkmark & $O(2TND)$ & $37.8$ &4& 2.26e10&\bm{$4.06\pm.052$}\\
\hline
\end{tabular}
\caption{Balance between consumption and synthesized quality of models. \textit{Spike-Driven} denotes the existence of solely AC operations. \textit{Param} represents the amount of parameters of models, \textit{Time Step} is total spike sequence time steps, and \textit{E(pJ)} represents the energy consumption calculated according to Table~\ref{energy consumption}. MOS represents the results of the LJSpeech.}
\label{trade-off}
\end{table*}

\subsection{Visualized Analysis}
\label{visualize}
\begin{figure}[h]
  \centering
  \subfigure[FastSpeech2]{
    \includegraphics[width=0.2\textwidth]{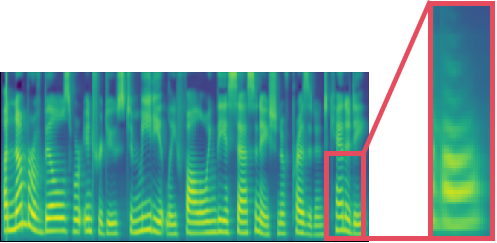}
    \label{subfig11}
  }
  \subfigure[SpikeVoice-ATTN]{
    \includegraphics[width=0.2\textwidth]{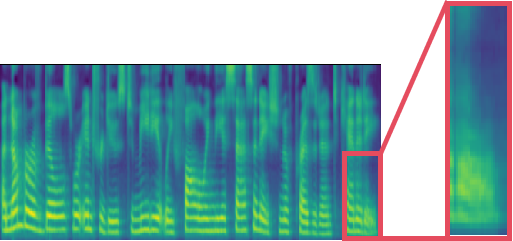}
    \label{subfig12}
  }
  \subfigure[SpikeVoice-SDSA]{
    \includegraphics[width=0.2\textwidth]{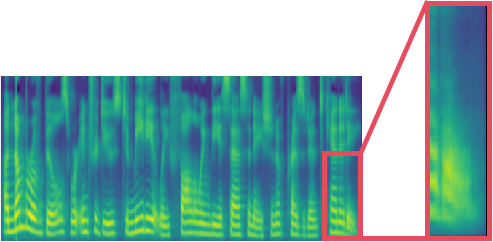}
    \label{subfig13}
  }
  \subfigure[SpikeVoice-STSA]{
    \includegraphics[width=0.2\textwidth]{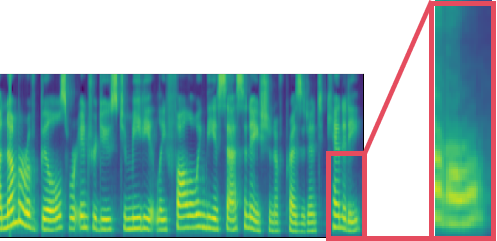}
    \label{subfig14}
  }
  \caption{Mel-Spectrograms visualization analysis on English single-speaker dataset LJSpeech.}
  \label{Mel demo}
\end{figure}

\textbf{Visualization of Mel-Spectrograms:} Speech synthesized by SpikeVoice exhibits less noise and is clearer compared to the SNN-based baselines, which is evident in Fig.\ref{Mel demo}. As shown in Fig. \ref{subfig12} and \ref{subfig13}, Mel-Spectrograms synthesized by the SNN baselines become blurry towards the end, losing fine details. In contrast, the Mel-Spectrograms in Fig.\ref{subfig14} synthesized by SpikeVoice with STSA exhibit minimal sacrifice of details as to ANN in \ref{subfig11} and remain notably clearer than those produced by SNN baselines. 

\begin{figure}[h]
  \centering
  \subfigure[STSA-layer1]{
    \includegraphics[width=0.18\textwidth]{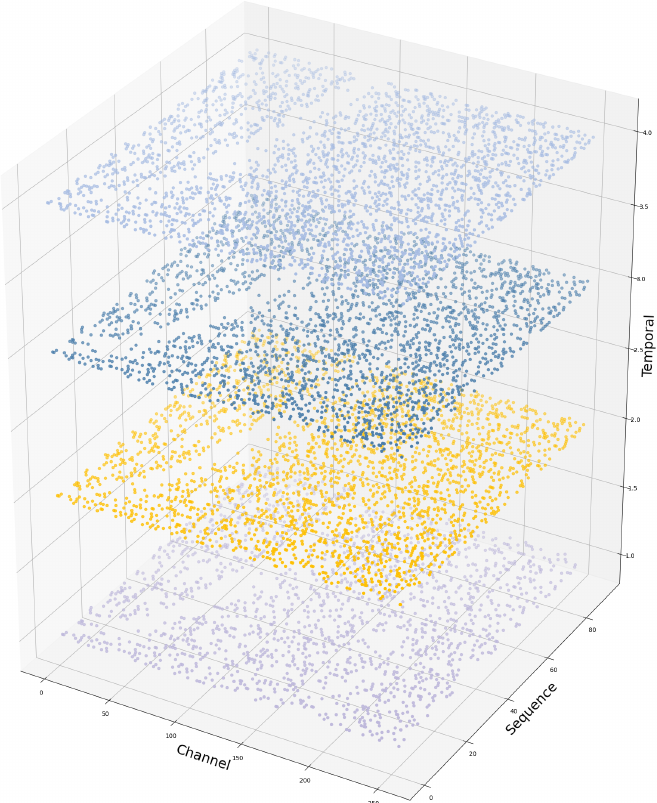}
    \label{subfig21}
  }
  \subfigure[STSA-layer4]{
    \includegraphics[width=0.18\textwidth]{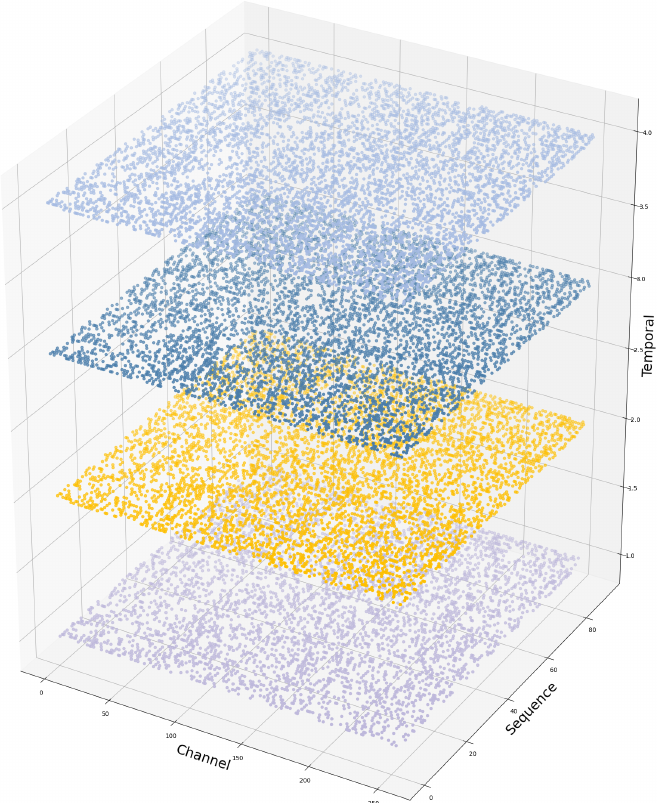}
    \label{subfig22}
  }
  \subfigure[Pitch Predictor]{
    \includegraphics[width=0.18\textwidth]{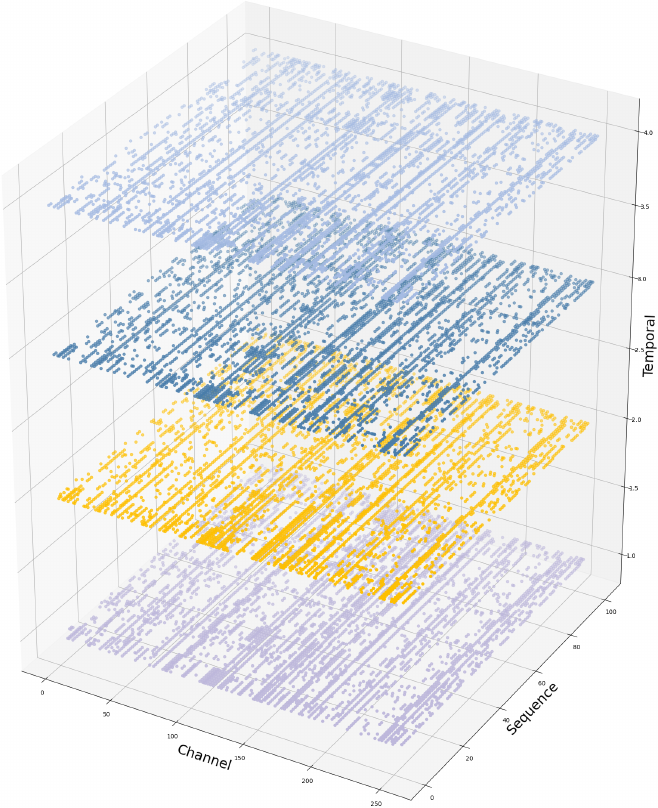}
    \label{subfig23}
  }
  \subfigure[Energy Predictor]{
    \includegraphics[width=0.18\textwidth]{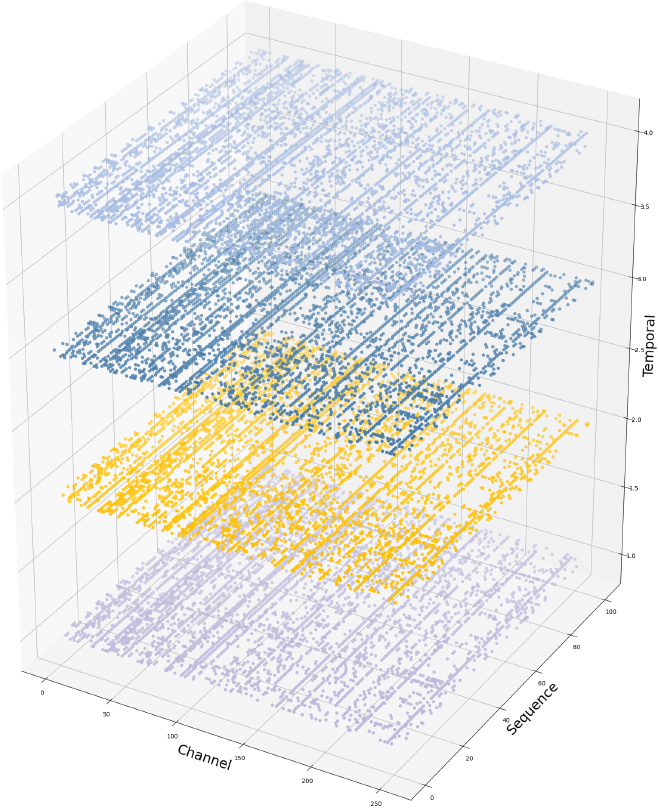}
    \label{subfig24}
  }
  \caption{Visualization of spike tensor. Fig.\ref{subfig21} and Fig.\ref{subfig22} are the spike patterns of STSA in the first layer and the fourth layer. \ref{subfig23} and \ref{subfig24} denote spike pattern for speech energy and speech pitch. Each dot depicts a fired event.}
  \label{Spike demo}
\end{figure}

\textbf{Visualization of Spike Patterns:} 
\label{visual_spike}
By visualizing spike tensors, more details of SpikeVoice can be observed. As the spike patterns of STSA depicted in Fig.\ref{subfig21} and Fig.\ref{subfig22}, each dot represents an event, the spike events in the lower layers are sparser, and as the network deepens, more information is incorporated, leading to denser spike events. Spike tensors that convey similar information exhibit similar spike patterns, while others reveal markedly different spike patterns.
Spike patterns of the energy and pitch predictors are displayed in Fig.\ref{subfig23} and Fig.\ref{subfig24}, different from the distribution of spike pattern in \ref{subfig21} and \ref{subfig22}, noticeable channel clustering phenomena can be observed in \ref{subfig23} and \ref{subfig24}. 

\subsection{Analysis of Balance between Consumption and Synthesized Speech Quality}
\label{balance}
Apart from its notable biological interpretability, one of the most prominent advantages of SNN lies in its energy efficiency. However, SNN's binary embedding within a finite time step results in some degree of performance decay. 
In Tab.\ref{trade-off}, we present the number of model parameters, time steps of binary embedding, and energy consumption. The term "Spike-Driven" refers to the existence of solely AC operations, and "MOS" here refers to the results on LJSpeech.

While SpikeVoice-STSA comes with a slight increase in the parameter, it takes only \textbf{10.5\%} energy consuming of ANN with $4$ time steps and achieves a better performance than SNN baselines. In contrast, SpikeVoice-SDSA exhibits noticeable performance degradation, while the energy consumption is \textbf{9.6\%} of ANN with an equivalent amount of parameters. Similarly, SpikeVoice-ATTN also results in an \textbf{88.1\%} reduction in energy consumption. It is worth to noting that when set time step to 1, the energy consumption of SpikeVoice-STSA can be merely \textbf{4.11\%} of ANN.
Hence, when considering both the quality of speech synthesis and energy consumption, SpikeVoice is a superior choice, offering significant energy savings with minimal performance sacrifice.

\section{Conclusion}
In this paper, we introduce SpikeVoice. To the best of our knowledge, it is the first TTS model that achieves high-quality speech synthesis within the SNN framework and for the first time endows SNN with the ability to "speak". Additionally, SpikeVoice is a spike-driven model with highly energy-efficient. In SpikeVoice, we propose STSA, which performs temporal-mixing in the spiking temporal dimension to address the issue of information invisibility at future time steps on the spiking temporal dimension caused by the serial nature of spiking neurons and thereby address the issue of "partial-time dependency".

We conducted experiments on both single-speaker and multi-speaker datasets in both Chinese and English. The results demonstrate that SpikeVoice achieves performance comparable to ANN models while consuming only $10.5\%$ of the energy required by ANN. Our successful practice proves the feasibility of TTS tasks within the SNN framework and offers an energy-saving solution for TTS tasks.

\section{Limitation}
The SpikeVoice within the SNN framework still has several limitations. Primarily, the binary embedding results in inevitably information lost from the input data, leading to a decline in performance. Secondly, due to the inherent sequential mechanism of LIF neurons, the training speed of SpikeVoice is slower than ANN. Finally, as analyzed in section \ref{visual_spike} with the layers deepen, the firing rate becomes progressively higher, which implies the potential for further reductions in energy consumption. In light of this, we present several prospective exploration directions that reduce information loss during the binary embedding process in SNN, lowering the firing rate in deep neural networks, and parallelization of spike neurons.

\section{Acknowledge}
This work was partially supported by National  Distinguished Young Scholars (62325603), and National Natural Science Foundation of China (62236009,U22A20103,62441606), and Beijing Natural Science Foundation for Distinguished Young Scholars (JQ21015), and CAAI-MindSpore Open Fund, developed on OpenI Community.

\bibliography{custom}

\begin{thebibliography}{51}
\expandafter\ifx\csname natexlab\endcsname\relax\def\natexlab#1{#1}\fi

\bibitem[{Abbott(1999)}]{IF1999}
Larry~F Abbott. 1999.
\newblock Lapicque’s introduction of the integrate-and-fire model neuron (1907).
\newblock \emph{Brain research bulletin}, 50(5-6):303--304.

\bibitem[{Anil et~al.(2023)Anil, Dai, Firat, Johnson, Lepikhin, Passos, Shakeri, Taropa, Bailey, Chen et~al.}]{palm2}
Rohan Anil, Andrew~M Dai, Orhan Firat, Melvin Johnson, Dmitry Lepikhin, Alexandre Passos, Siamak Shakeri, Emanuel Taropa, Paige Bailey, Zhifeng Chen, et~al. 2023.
\newblock Palm 2 technical report.
\newblock \emph{arXiv preprint arXiv:2305.10403}.

\bibitem[{Baevski et~al.(2020)Baevski, Zhou, Mohamed, and Auli}]{wav2vec2}
Alexei Baevski, Yuhao Zhou, Abdelrahman Mohamed, and Michael Auli. 2020.
\newblock wav2vec 2.0: A framework for self-supervised learning of speech representations.
\newblock \emph{Advances in neural information processing systems}, 33:12449--12460.

\bibitem[{Brown et~al.(2020)Brown, Mann, Ryder, Subbiah, Kaplan, Dhariwal, Neelakantan, Shyam, Sastry, Askell et~al.}]{gpt}
Tom Brown, Benjamin Mann, Nick Ryder, Melanie Subbiah, Jared~D Kaplan, Prafulla Dhariwal, Arvind Neelakantan, Pranav Shyam, Girish Sastry, Amanda Askell, et~al. 2020.
\newblock Language models are few-shot learners.
\newblock \emph{Advances in neural information processing systems}, 33:1877--1901.

\bibitem[{Bu et~al.(2021)Bu, Fang, Ding, DAI, Yu, and Huang}]{optimal}
Tong Bu, Wei Fang, Jianhao Ding, PENGLIN DAI, Zhaofei Yu, and Tiejun Huang. 2021.
\newblock Optimal ann-snn conversion for high-accuracy and ultra-low-latency spiking neural networks.
\newblock In \emph{International Conference on Learning Representations}.

\bibitem[{Cao et~al.(2015)Cao, Chen, and Khosla}]{spiking_energy}
Yongqiang Cao, Yang Chen, and Deepak Khosla. 2015.
\newblock Spiking deep convolutional neural networks for energy-efficient object recognition.
\newblock \emph{International Journal of Computer Vision}, 113:54--66.

\bibitem[{Carion et~al.(2020)Carion, Massa, Synnaeve, Usunier, Kirillov, and Zagoruyko}]{dert}
Nicolas Carion, Francisco Massa, Gabriel Synnaeve, Nicolas Usunier, Alexander Kirillov, and Sergey Zagoruyko. 2020.
\newblock End-to-end object detection with transformers.
\newblock In \emph{European conference on computer vision}, pages 213--229.

\bibitem[{Deng and Gu(2020)}]{deng2021optimal}
Shikuang Deng and Shi Gu. 2020.
\newblock Optimal conversion of conventional artificial neural networks to spiking neural networks.
\newblock In \emph{International Conference on Learning Representations}.

\bibitem[{Devlin et~al.(2019)Devlin, Chang, Lee, and Toutanova}]{bert}
Jacob Devlin, Ming-Wei Chang, Kenton Lee, and Kristina Toutanova. 2019.
\newblock Bert: Pre-training of deep bidirectional transformers for language understanding.
\newblock In \emph{Proceedings of NAACL-HLT}.

\bibitem[{Duan et~al.(2022)Duan, Ding, Chen, Yu, and Huang}]{duan2022temporal}
Chaoteng Duan, Jianhao Ding, Shiyan Chen, Zhaofei Yu, and Tiejun Huang. 2022.
\newblock Temporal effective batch normalization in spiking neural networks.
\newblock \emph{Advances in Neural Information Processing Systems}, 35:34377--34390.

\bibitem[{Fang et~al.(2023{\natexlab{a}})Fang, Chen, Ding, Yu, Masquelier, Chen, Huang, Zhou, Li, and Tian}]{fang2023spikingjelly}
Wei Fang, Yanqi Chen, Jianhao Ding, Zhaofei Yu, Timoth{\'e}e Masquelier, Ding Chen, Liwei Huang, Huihui Zhou, Guoqi Li, and Yonghong Tian. 2023{\natexlab{a}}.
\newblock Spikingjelly: An open-source machine learning infrastructure platform for spike-based intelligence.
\newblock \emph{Science Advances}, 9(40):eadi1480.

\bibitem[{Fang et~al.(2023{\natexlab{b}})Fang, Yu, Zhou, Chen, Chen, Ma, Masquelier, and Tian}]{psn}
Wei Fang, Zhaofei Yu, Zhaokun Zhou, Ding Chen, Yanqi Chen, Zhengyu Ma, Timoth{\'e}e Masquelier, and Yonghong Tian. 2023{\natexlab{b}}.
\newblock Parallel spiking neurons with high efficiency and ability to learn long-term dependencies.
\newblock In \emph{Thirty-seventh Conference on Neural Information Processing Systems}.

\bibitem[{Han et~al.(2020)Han, Srinivasan, and Roy}]{han2020rmp}
Bing Han, Gopalakrishnan Srinivasan, and Kaushik Roy. 2020.
\newblock Rmp-snn: Residual membrane potential neuron for enabling deeper high-accuracy and low-latency spiking neural network.
\newblock In \emph{Proceedings of the IEEE/CVF conference on computer vision and pattern recognition}, pages 13558--13567.

\bibitem[{Hochreiter and Schmidhuber(1997)}]{rnn}
Sepp Hochreiter and J{\"u}rgen Schmidhuber. 1997.
\newblock Long short-term memory.
\newblock \emph{Neural computation}, 9(8):1735--1780.

\bibitem[{Hodgkin and Huxley(1952)}]{HH}
Alan~L Hodgkin and Andrew~F Huxley. 1952.
\newblock A quantitative description of membrane current and its application to conduction and excitation in nerve.
\newblock \emph{The Journal of physiology}, 117(4):500.

\bibitem[{Hsu et~al.(2021)Hsu, Bolte, Tsai, Lakhotia, Salakhutdinov, and Mohamed}]{hubert}
Wei-Ning Hsu, Benjamin Bolte, Yao-Hung~Hubert Tsai, Kushal Lakhotia, Ruslan Salakhutdinov, and Abdelrahman Mohamed. 2021.
\newblock Hubert: Self-supervised speech representation learning by masked prediction of hidden units.
\newblock \emph{IEEE/ACM Transactions on Audio, Speech, and Language Processing}, 29:3451--3460.

\bibitem[{Ito and Johnson(2017)}]{ljspeech}
Keith Ito and Linda Johnson. 2017.
\newblock The lj speech dataset.
\newblock \url{https://keithito.com/LJ-Speech-Dataset/}.

\bibitem[{Kong et~al.(2020)Kong, Kim, and Bae}]{hifigan}
Jungil Kong, Jaehyeon Kim, and Jaekyoung Bae. 2020.
\newblock Hifi-gan: Generative adversarial networks for efficient and high fidelity speech synthesis.
\newblock \emph{Advances in Neural Information Processing Systems}, 33:17022--17033.

\bibitem[{Li et~al.(2023{\natexlab{a}})Li, Li, Savarese, and Hoi}]{blip2}
Junnan Li, Dongxu Li, Silvio Savarese, and Steven Hoi. 2023{\natexlab{a}}.
\newblock Blip-2: Bootstrapping language-image pre-training with frozen image encoders and large language models.
\newblock In \emph{International conference on machine learning}.

\bibitem[{Li et~al.(2019)Li, Liu, Liu, Zhao, and Liu}]{transformer-tts}
Naihan Li, Shujie Liu, Yanqing Liu, Sheng Zhao, and Ming Liu. 2019.
\newblock Neural speech synthesis with transformer network.
\newblock In \emph{Proceedings of the AAAI conference on artificial intelligence}, volume~33, pages 6706--6713.

\bibitem[{Li et~al.(2023{\natexlab{b}})Li, Liu, Lv, Xu, Zhang, Wu, Zheng, and Huang}]{spikeclip}
Tianlong Li, Wenhao Liu, Changze Lv, Jianhan Xu, Cenyuan Zhang, Muling Wu, Xiaoqing Zheng, and Xuanjing Huang. 2023{\natexlab{b}}.
\newblock Spikeclip: A contrastive language-image pretrained spiking neural network.
\newblock \emph{arXiv preprint arXiv:2310.06488}.

\bibitem[{Liu et~al.(2021)Liu, Lin, Cao, Hu, Wei, Zhang, Lin, and Guo}]{swin}
Ze~Liu, Yutong Lin, Yue Cao, Han Hu, Yixuan Wei, Zheng Zhang, Stephen Lin, and Baining Guo. 2021.
\newblock Swin transformer: Hierarchical vision transformer using shifted windows.
\newblock In \emph{Proceedings of the IEEE/CVF international conference on computer vision}, pages 10012--10022.

\bibitem[{Lv et~al.(2023)Lv, Li, Xu, Gu, Ling, Zhang, Zheng, and Huang}]{spikebert}
Changze Lv, Tianlong Li, Jianhan Xu, Chenxi Gu, Zixuan Ling, Cenyuan Zhang, Xiaoqing Zheng, and Xuanjing Huang. 2023.
\newblock Spikebert: A language spikformer trained with two-stage knowledge distillation from bert.
\newblock \emph{arXiv preprint arXiv:2308.15122}.

\bibitem[{Lv et~al.(2022)Lv, Xu, and Zheng}]{spike_text}
Changze Lv, Jianhan Xu, and Xiaoqing Zheng. 2022.
\newblock Spiking convolutional neural networks for text classification.
\newblock In \emph{The Eleventh International Conference on Learning Representations}.

\bibitem[{Maass(1997)}]{spiking_first}
Wolfgang Maass. 1997.
\newblock Networks of spiking neurons: the third generation of neural network models.
\newblock \emph{Neural networks}, 10(9):1659--1671.

\bibitem[{Mittag et~al.(2021)Mittag, Naderi, Chehadi, and M{\"o}ller}]{nisqa}
Gabriel Mittag, Babak Naderi, Assmaa Chehadi, and Sebastian M{\"o}ller. 2021.
\newblock Nisqa: A deep cnn-self-attention model for multidimensional speech quality prediction with crowdsourced datasets.
\newblock \emph{arXiv preprint arXiv:2104.09494}.

\bibitem[{Mueller et~al.(2021)Mueller, Studenyak, Auge, and Knoll}]{convert}
Etienne Mueller, Viktor Studenyak, Daniel Auge, and Alois Knoll. 2021.
\newblock Spiking transformer networks: A rate coded approach for processing sequential data.
\newblock In \emph{2021 7th International Conference on Systems and Informatics (ICSAI)}, pages 1--5.

\bibitem[{OpenAI(2023)}]{gpt4}
OpenAI. 2023.
\newblock \href {https://api.semanticscholar.org/CorpusID:257532815} {Gpt-4 technical report}.
\newblock \emph{ArXiv}, abs/2303.08774.

\bibitem[{Radford et~al.(2021)Radford, Kim, Hallacy, Ramesh, Goh, Agarwal, Sastry, Askell, Mishkin, Clark et~al.}]{clip}
Alec Radford, Jong~Wook Kim, Chris Hallacy, Aditya Ramesh, Gabriel Goh, Sandhini Agarwal, Girish Sastry, Amanda Askell, Pamela Mishkin, Jack Clark, et~al. 2021.
\newblock Learning transferable visual models from natural language supervision.
\newblock In \emph{International conference on machine learning}, pages 8748--8763.

\bibitem[{Radford et~al.(2023)Radford, Kim, Xu, Brockman, McLeavey, and Sutskever}]{whisper}
Alec Radford, Jong~Wook Kim, Tao Xu, Greg Brockman, Christine McLeavey, and Ilya Sutskever. 2023.
\newblock Robust speech recognition via large-scale weak supervision.
\newblock In \emph{International Conference on Machine Learning}, pages 28492--28518.

\bibitem[{Rajagopal et~al.(2023)Rajagopal, Karthick, Meenalochini, and Kalaichelvi}]{image_detect2}
RKPMTKR Rajagopal, R~Karthick, P~Meenalochini, and T~Kalaichelvi. 2023.
\newblock Deep convolutional spiking neural network optimized with arithmetic optimization algorithm for lung disease detection using chest x-ray images.
\newblock \emph{Biomedical Signal Processing and Control}, 79:104197.

\bibitem[{Ren et~al.(2020)Ren, Hu, Tan, Qin, Zhao, Zhao, and Liu}]{fastspeech2}
Yi~Ren, Chenxu Hu, Xu~Tan, Tao Qin, Sheng Zhao, Zhou Zhao, and Tie-Yan Liu. 2020.
\newblock Fastspeech 2: Fast and high-quality end-to-end text to speech.
\newblock In \emph{International Conference on Learning Representations}.

\bibitem[{Ren et~al.(2019)Ren, Ruan, Tan, Qin, Zhao, Zhao, and Liu}]{fastspeech}
Yi~Ren, Yangjun Ruan, Xu~Tan, Tao Qin, Sheng Zhao, Zhou Zhao, and Tie-Yan Liu. 2019.
\newblock Fastspeech: Fast, robust and controllable text to speech.
\newblock \emph{Advances in neural information processing systems}, 32.

\bibitem[{Shen et~al.(2018)Shen, Pang, Weiss, Schuster, Jaitly, Yang, Chen, Zhang, Wang, Skerrv-Ryan et~al.}]{tacotron2}
Jonathan Shen, Ruoming Pang, Ron~J Weiss, Mike Schuster, Navdeep Jaitly, Zongheng Yang, Zhifeng Chen, Yu~Zhang, Yuxuan Wang, Rj~Skerrv-Ryan, et~al. 2018.
\newblock Natural tts synthesis by conditioning wavenet on mel spectrogram predictions.
\newblock In \emph{2018 IEEE international conference on acoustics, speech and signal processing (ICASSP)}, pages 4779--4783.

\bibitem[{Shrestha and Orchard(2018)}]{shrestha2018slayer}
Sumit~B Shrestha and Garrick Orchard. 2018.
\newblock Slayer: Spike layer error reassignment in time.
\newblock \emph{Advances in neural information processing systems}, 31.

\bibitem[{Sun et~al.(2023)Sun, Yu, Cui, Zhang, Zhang, Wang, Gao, Liu, Huang, and Wang}]{emu}
Quan Sun, Qiying Yu, Yufeng Cui, Fan Zhang, Xiaosong Zhang, Yueze Wang, Hongcheng Gao, Jingjing Liu, Tiejun Huang, and Xinlong Wang. 2023.
\newblock Generative pretraining in multimodality.
\newblock \emph{arXiv preprint arXiv:2307.05222}.

\bibitem[{Touvron et~al.(2023)Touvron, Martin, Stone, Albert, Almahairi, Babaei, Bashlykov, Batra, Bhargava, Bhosale et~al.}]{llama2}
Hugo Touvron, Louis Martin, Kevin Stone, Peter Albert, Amjad Almahairi, Yasmine Babaei, Nikolay Bashlykov, Soumya Batra, Prajjwal Bhargava, Shruti Bhosale, et~al. 2023.
\newblock Llama 2: Open foundation and fine-tuned chat models.
\newblock \emph{arXiv preprint arXiv:2307.09288}.

\bibitem[{Vaswani et~al.(2017)Vaswani, Shazeer, Parmar, Uszkoreit, Jones, Gomez, Kaiser, and Polosukhin}]{transformer}
Ashish Vaswani, Noam Shazeer, Niki Parmar, Jakob Uszkoreit, Llion Jones, Aidan~N Gomez, {\L}ukasz Kaiser, and Illia Polosukhin. 2017.
\newblock Attention is all you need.
\newblock \emph{Advances in neural information processing systems}, 30.

\bibitem[{Wang et~al.(2023)Wang, Zhang, Han, Wang, Zhang, and Xu}]{spike_speech2}
Qingyu Wang, Tielin Zhang, Minglun Han, Yi~Wang, Duzhen Zhang, and Bo~Xu. 2023.
\newblock Complex dynamic neurons improved spiking transformer network for efficient automatic speech recognition.
\newblock In \emph{Proceedings of the AAAI Conference on Artificial Intelligence}, volume~37, pages 102--109.

\bibitem[{Wu et~al.(2020)Wu, Y{\i}lmaz, Zhang, Li, and Tan}]{spike_speech1}
Jibin Wu, Emre Y{\i}lmaz, Malu Zhang, Haizhou Li, and Kay~Chen Tan. 2020.
\newblock Deep spiking neural networks for large vocabulary automatic speech recognition.
\newblock \emph{Frontiers in neuroscience}, 14:199.

\bibitem[{Wu et~al.(2018{\natexlab{a}})Wu, Deng, Li, Zhu, and Shi}]{wu2018spatio}
Yujie Wu, Lei Deng, Guoqi Li, Jun Zhu, and Luping Shi. 2018{\natexlab{a}}.
\newblock Spatio-temporal backpropagation for training high-performance spiking neural networks.
\newblock \emph{Frontiers in neuroscience}, 12:331.

\bibitem[{Wu et~al.(2018{\natexlab{b}})Wu, Deng, Li, Zhu, and Shi}]{stbp}
Yujie Wu, Lei Deng, Guoqi Li, Jun Zhu, and Luping Shi. 2018{\natexlab{b}}.
\newblock Spatio-temporal backpropagation for training high-performance spiking neural networks.
\newblock \emph{Frontiers in neuroscience}, 12:331.

\bibitem[{Yao et~al.(2024{\natexlab{a}})Yao, Hu, Hu, Xu, Zhou, Tian, XU, and Li}]{spikedrivenv2}
Man Yao, JiaKui Hu, Tianxiang Hu, Yifan Xu, Zhaokun Zhou, Yonghong Tian, Bo~XU, and Guoqi Li. 2024{\natexlab{a}}.
\newblock Spike-driven transformer v2: Meta spiking neural network architecture inspiring the design of next-generation neuromorphic chips.
\newblock In \emph{The Twelfth International Conference on Learning Representations}.

\bibitem[{Yao et~al.(2023{\natexlab{a}})Yao, Hu, Zhou, Yuan, Tian, Bo, and Li}]{spike-driven}
Man Yao, JiaKui Hu, Zhaokun Zhou, Li~Yuan, Yonghong Tian, XU~Bo, and Guoqi Li. 2023{\natexlab{a}}.
\newblock Spike-driven transformer.
\newblock In \emph{Thirty-seventh Conference on Neural Information Processing Systems}.

\bibitem[{Yao et~al.(2024{\natexlab{b}})Yao, Richter, Zhao, Qiao, Xing, Wang, Hu, Fang, Demirci, De~Marchi, Deng, Yan, Nielsen, Sheik, Wu, Tian, Xu, and Li}]{Speck}
Man Yao, Ole Richter, Guangshe Zhao, Ning Qiao, Yannan Xing, Dingheng Wang, Tianxiang Hu, Wei Fang, Tugba Demirci, Michele De~Marchi, Lei Deng, Tianyi Yan, Carsten Nielsen, Sadique Sheik, Chenxi Wu, Yonghong Tian, Bo~Xu, and Guoqi Li. 2024{\natexlab{b}}.
\newblock Spike-based dynamic computing with asynchronous sensing-computing neuromorphic chip.
\newblock \emph{Nature Communications}, 15(1):4464.

\bibitem[{Yao et~al.(2023{\natexlab{b}})Yao, Zhao, Zhang, Hu, Deng, Tian, Xu, and Li}]{AttentionSNN}
Man Yao, Guangshe Zhao, Hengyu Zhang, Yifan Hu, Lei Deng, Yonghong Tian, Bo~Xu, and Guoqi Li. 2023{\natexlab{b}}.
\newblock Attention spiking neural networks.
\newblock \emph{IEEE Transactions on Pattern Analysis and Machine Intelligence}.

\bibitem[{Yao et~al.(2015)Yao, Hui, Xin, Shaoji, and Ming}]{AISHELL-3}
Shi Yao, Bu~Hui, Xu~Xin, Zhang Shaoji, and Li~Ming. 2015.
\newblock \href {https://arxiv.org/abs/2010.11567} {Aishell-3: A multi-speaker mandarin tts corpus and the baselines}.

\bibitem[{Zen et~al.(2019)Zen, Dang, Clark, Zhang, Weiss, Jia, Chen, and Wu}]{libritts}
Heiga Zen, Viet Dang, Rob Clark, Yu~Zhang, Ron~J Weiss, Ye~Jia, Zhifeng Chen, and Yonghui Wu. 2019.
\newblock Libritts: A corpus derived from librispeech for text-to-speech.
\newblock \emph{arXiv preprint arXiv:1904.02882}.

\bibitem[{Zhao et~al.(2021)Zhao, Zhang, Yan, Qiu, Yao, Tian, Zhu, and Cao}]{image_detect1}
Jianqing Zhao, Xiaohu Zhang, Jiawei Yan, Xiaolei Qiu, Xia Yao, Yongchao Tian, Yan Zhu, and Weixing Cao. 2021.
\newblock A wheat spike detection method in uav images based on improved yolov5.
\newblock \emph{Remote Sensing}, 13(16):3095.

\bibitem[{Zhou et~al.(2022)Zhou, Zhu, He, Wang, Shuicheng, Tian, and Yuan}]{spikeformer}
Zhaokun Zhou, Yuesheng Zhu, Chao He, Yaowei Wang, YAN Shuicheng, Yonghong Tian, and Li~Yuan. 2022.
\newblock Spikformer: When spiking neural network meets transformer.
\newblock In \emph{The Eleventh International Conference on Learning Representations}.

\bibitem[{Zhu et~al.(2023)Zhu, Zhao, Li, and Eshraghian}]{spikegpt}
Rui-Jie Zhu, Qihang Zhao, Guoqi Li, and Jason~K Eshraghian. 2023.
\newblock Spikegpt: Generative pre-trained language model with spiking neural networks.
\newblock \emph{arXiv preprint arXiv:2302.13939}.

\end{thebibliography}

\appendix

\section{Firing Rate of SpikeVoice}
In Tab.\ref{fr_encoder}, Tab.\ref{fr_vd}, and Tab.\ref{fr_decoder}, we respectively present the spike firing rates of Spiking Phoneme Encoder, Spiking Variance Adapter, and Spiking Mel Decoder.
\label{sec:appendix}
\begin{table*}
\setlength{\tabcolsep}{0.36cm}
\centering
\begin{tabular}{c|c|cccc|c}
\hline
\multicolumn{7}{c}{Spiking Phoneme Encoder}\\
\hline
& & Layer1 & Layer2 & Layer3 & Layer4& AVG\\
\hline
\multirow{4}{*}{\makecell[c]{Spiking Sequential Attention}}  & Q & 0.19 & 0.18 & 0.19&0.2&0.19 \\
& K & 0.04 & 0.04 & 0.05&0.07&0.05 \\
& V & 0.04 & 0.04 & 0.05&0.07&0.05 \\
& Linear & 0.05 & 0.05 & 0.06 & 0.09&0.06\\
\hline
\multirow{4}{*}{\makecell[c]{Spiking Temporal Attention}}  & Q & 0.04 & 0.02 & 0.02&0.03&0.03 \\
& K & 0.05 & 0.02 & 0.03&0.04&0.04 \\
& V & 0.05 & 0.03 & 0.03&0.04&0.04 \\
& Linear & 0.01 & 0.01 & 0.01&0.02&0.01\\
\hline
\multirow{2}{*}{\makecell[c]{Spiking FeedForward}} & Conv1 & 0.07 & 0.10 & 0.13&0.15&0.11\\
& Conv2 & 0.12 & 0.10 & 0.12&0.17&0.13\\
\hline
\end{tabular}
\caption{Spike Firing Rates in Spiking Phoneme Encoder of SpikeVoice on LJSpeech dataset. The spike firing rate refers to the proportion of elements in the spike tensor that have an activation value of 1, with the value of other elements being 0.}
\label{fr_encoder}
\end{table*}

\begin{table*}
\setlength{\tabcolsep}{0.4cm}
\centering
\begin{tabular}{c|ccc|c}
\hline
\multicolumn{5}{c}{Spiking Variance Adapter}\\
\hline
 & FR\_Conv1 & FR\_Conv2 & FR\_Conv3 &  AVG\\
\hline
Duration Predictor  & 0.23 & 0.29 & 0.24&0.25 \\
Energy Predictor  & 0.27 & 0.31 & 0.32&0.30 \\
Pitch Predictor  & 0.23 & 0.38 & 0.30&0.30 \\
\hline
\end{tabular}
\caption{Spike Firing Rates in Spiking Variance Adapter of SpikeVoice on LJSpeech dataset. "FR\_Conv1", "FR\_Conv2" and "FR\_Conv3" in the SpikeVoice refer to the firing rate in Conv1, Conv2, and Conv3 of the Predictors respectively. }
\label{fr_vd}
\end{table*}

\begin{table*}
\setlength{\tabcolsep}{0.16cm}
\centering
\begin{tabular}{c|c|cccccc|c}
\hline
\multicolumn{9}{c}{Spiking Mel Decoder}\\
\hline
& & Layer1 & Layer2 & Layer3 & Layer4& Layer5 & Layer6& AVG\\
\hline
\multirow{4}{*}{\makecell[c]{Spiking Sequential Attention}}  & Q & 0.16 & 0.17 & 0.18&0.21&0.24&0.31&0.21 \\
& K & 0.03 & 0.04 & 0.04&0.05&0.05&0.04&0.04 \\
& V & 0.03 & 0.04 & 0.04&0.04&0.05&0.05&0.04 \\
& Linear & 0.03 & 0.05 & 0.06&0.07&0.8&0.11&0.07\\
\hline
\multirow{4}{*}{\makecell[c]{Spiking Temporal Attention}}  & Q & 0.14 & 0.13 & 0.14 & 0.13&0.13&0.13&0.13\\
& K & 0.24 & 0.20 & 0.18&0.18&0.19&0.22&0.20 \\
& V & 0.24 & 0.20 & 0.18&0.18&0.19&0.21&0.20 \\
& Linear & 0.02 & 0.02 & 0.03&0.03&0.03&0.04&0.03\\
\hline
\multirow{2}{*}{\makecell[c]{Spiking FeedForward}} & Conv1 & 0.12 & 0.13 & 0.13&0.13&0.12&0.19&0.14\\
& Conv2 & 0.10 & 0.13 & 0.14&0.15&0.16&0.22&0.15\\
\hline
\end{tabular}
\caption{Spike Firing Rates in Spiking Mel Decoder of SpikeVoice on LJSpeech dataset. The spike firing rate refers to the proportion of elements in the spike tensor that have an activation value of 1, with the value of other elements being 0.}
\label{fr_decoder}
\end{table*}

\section{Examples of Spike Patterns}
\begin{figure*}
  \centering
  \subfigure[STSA-encoder1]{
    \includegraphics[width=0.24\textwidth]{IMAGE/encoder_0.pdf}
    \label{encoder1}
  }
  \subfigure[STSA-encoder2]{
    \includegraphics[width=0.24\textwidth]{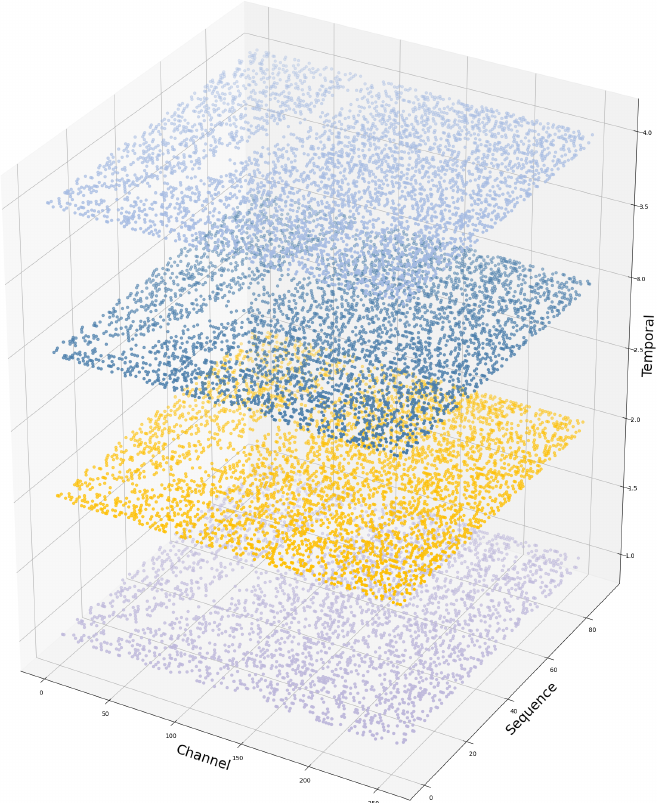}
    \label{encoder2}
  }
  \subfigure[STSA-encoder3]{
    \includegraphics[width=0.24\textwidth]{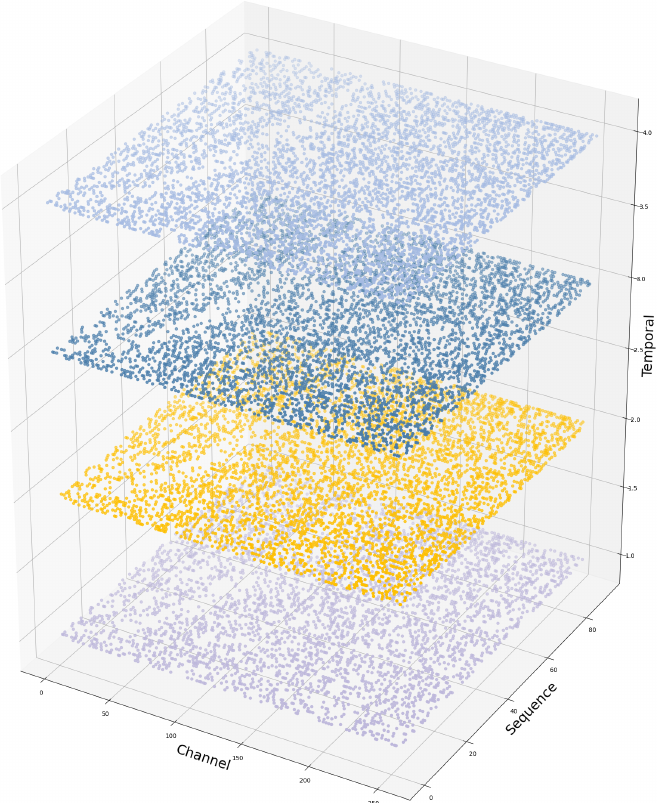}
    \label{encoder3}
  }
  \subfigure[STSA-encoder4]{
    \includegraphics[width=0.24\textwidth]{IMAGE/encoder_3.pdf}
    \label{encoder4}
  }
  \subfigure[Energy Predictor]{
    \includegraphics[width=0.24\textwidth]{IMAGE/out_e.pdf}
    \label{e_out}
  }
  \subfigure[Pitch Predictor]{
    \includegraphics[width=0.24\textwidth]{IMAGE/out_p.pdf}
    \label{p_out}
  }
  \subfigure[STSA-decoder1]{
    \includegraphics[width=0.24\textwidth]{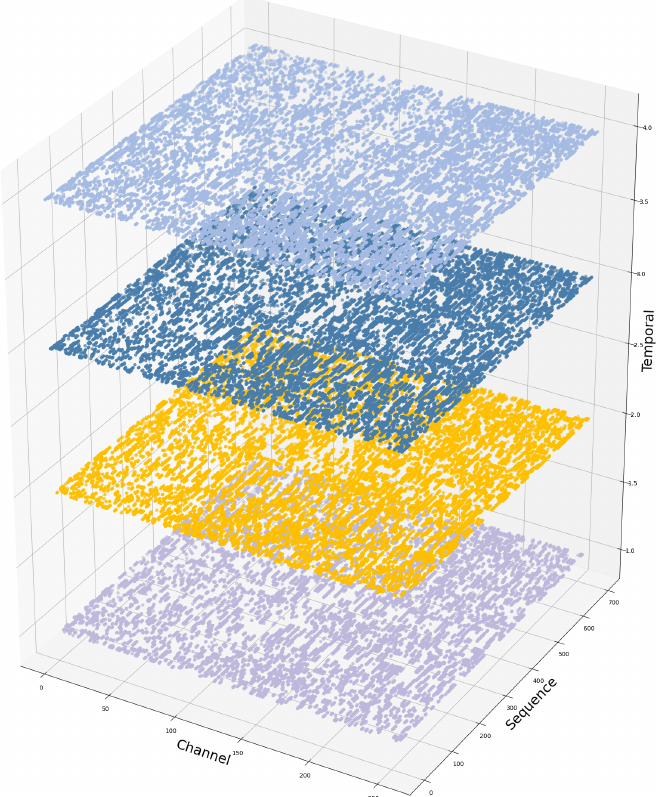}
    \label{decoder1}
  }
  \subfigure[STSA-decoder2]{
    \includegraphics[width=0.24\textwidth]{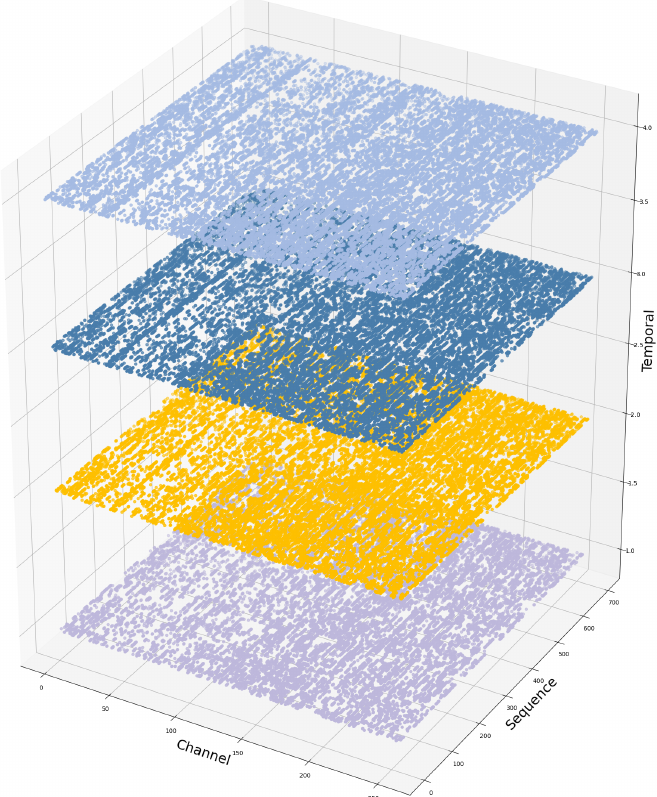}
    \label{decoder2}
  }
  \subfigure[STSA-decoder3]{
    \includegraphics[width=0.24\textwidth]{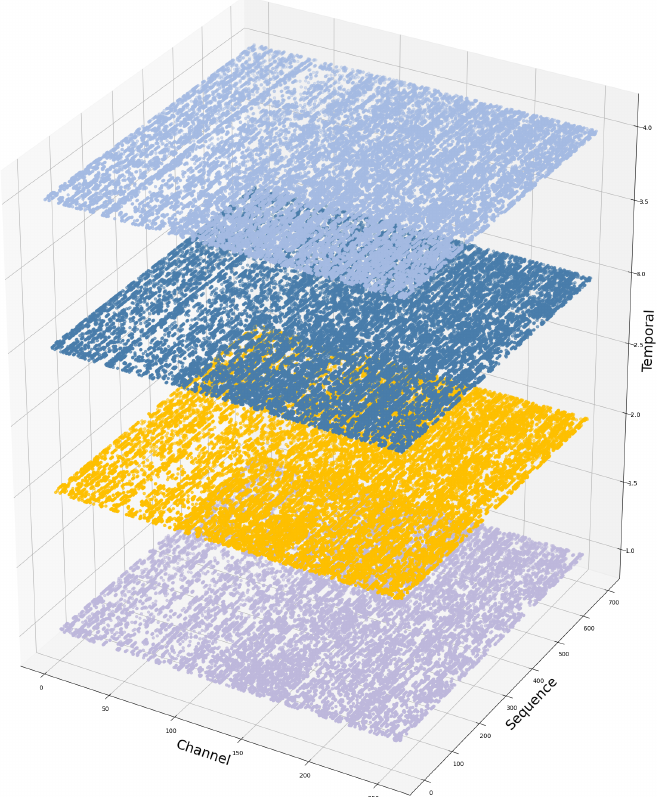}
    \label{decoder3}
  }
  \subfigure[STSA-decoder4]{
    \includegraphics[width=0.24\textwidth]{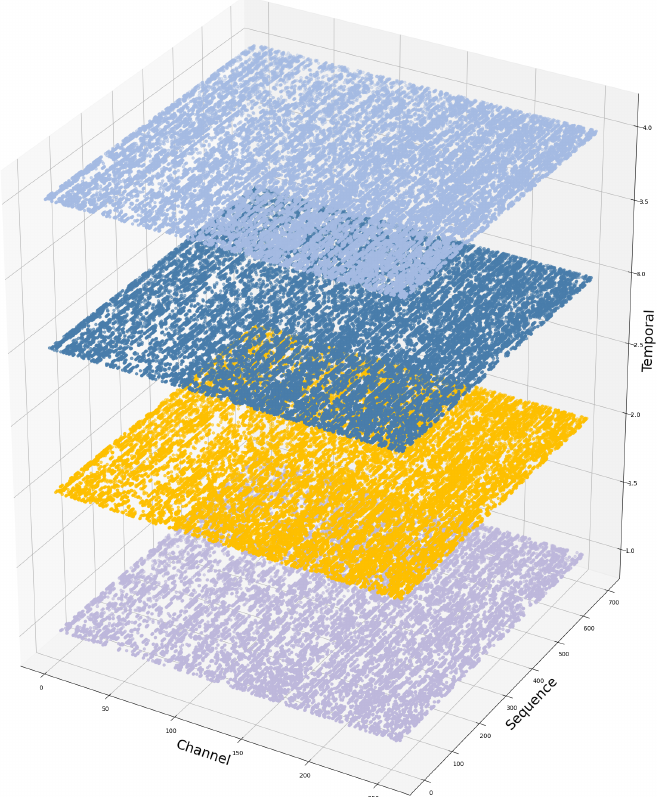}
    \label{decoder4}
  }
  \subfigure[STSA-decoder5]{
    \includegraphics[width=0.24\textwidth]{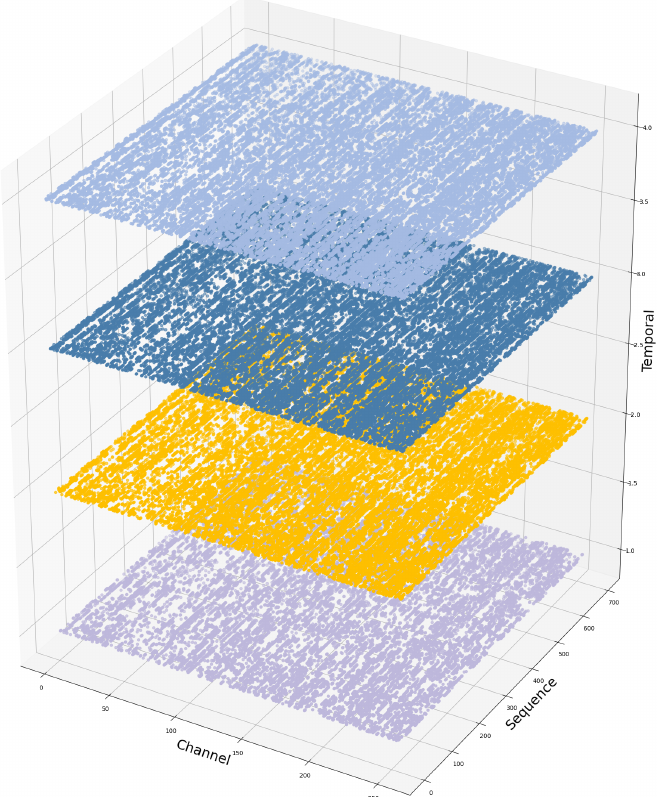}
    \label{decoder5}
  }
  \subfigure[STSA-decoder6]{
    \includegraphics[width=0.24\textwidth]{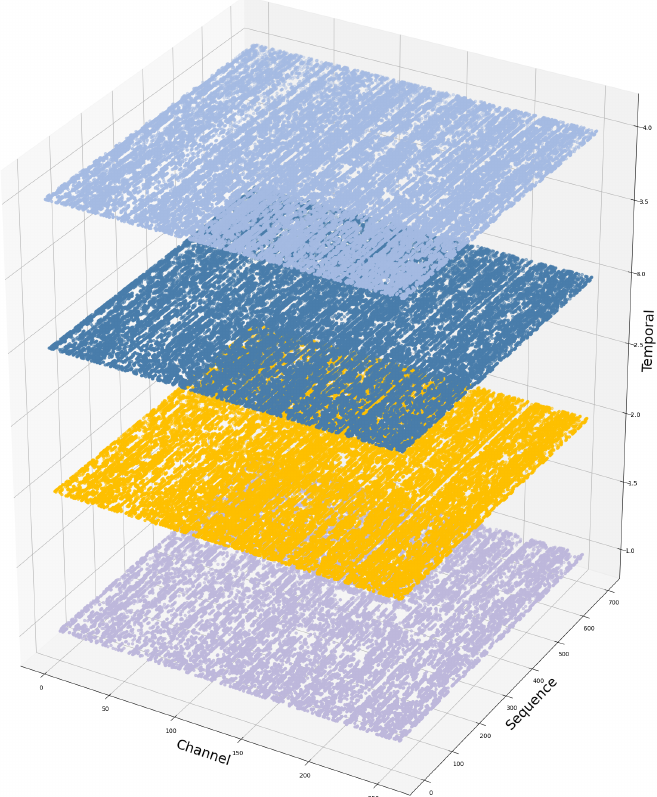}
    \label{decoder6}
  }
  \caption{Visualization of spike tensor in the SpikeVoice. Figures in \ref{encoder1},\ref{encoder2},\ref{encoder3},\ref{encoder4} are the spike pattern of STSA in Spiking Phoneme Encoder. \ref{e_out} and \ref{p_out} denote spike pattern for speech energy and speech pitch. Fig.\ref{decoder1} to \ref{decoder6} are the spike pattern of STSA in Spiking Mel Decoder.}
  \label{Spike pattern demo}
\end{figure*}

In Fig.\ref{Spike pattern demo} we present the spike patterns of STSA and also the spike patterns of Pitch Predictor and Energy Predictor.

\section{Examples of Mel-Spectrograms}
\begin{figure*}[h]
  \centering
  \subfigure[Mel-Spectrograms of LJSpeech]{
    \includegraphics[width=0.24\textwidth]{IMAGE/ljspeech_ann.png}
    \label{subfig11_fl}
    \includegraphics[width=0.24\textwidth]{IMAGE/ljspeech_attn.png}
    \label{subfig12_fl}
    \includegraphics[width=0.24\textwidth]{IMAGE/ljspeech_sdsa.png}
    \label{subfig13_fl}
    \includegraphics[width=0.24\textwidth]{IMAGE/ljspeech_sdsa_st.png}
    \label{subfig14_fl}
  }
  \medskip
  \vspace{0.5cm}
  \subfigure[Mel-Spectrograms of Baker]{
    \includegraphics[width=0.24\textwidth]{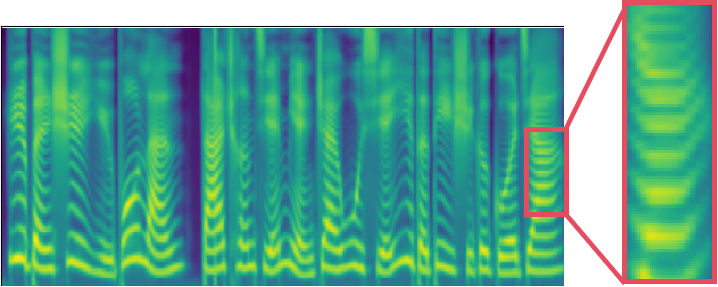}
    \label{subfig21_fl}
    \includegraphics[width=0.24\textwidth]{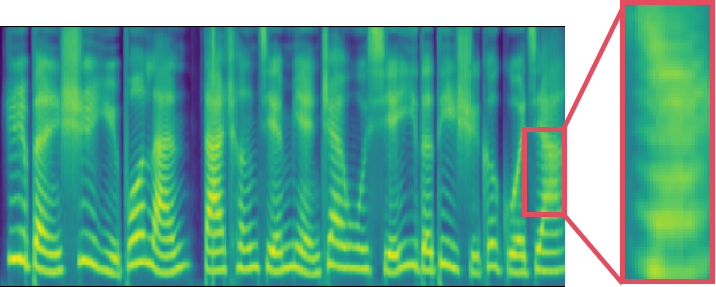}
    \label{subfig22_fl}
    \includegraphics[width=0.24\textwidth]{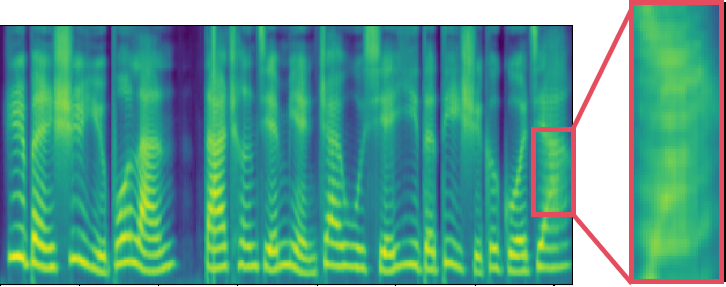}
    \label{subfig23_fl}
    \includegraphics[width=0.24\textwidth]{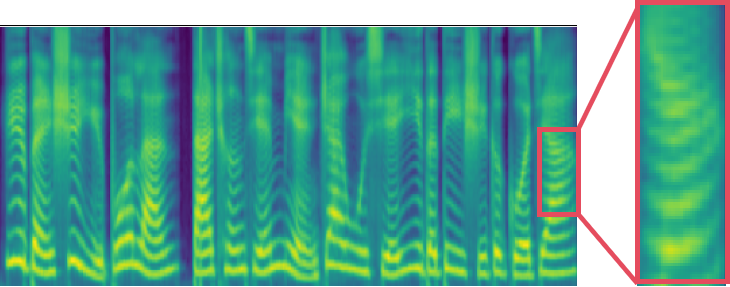}
    \label{subfig24_fl}
  }
  \medskip
  \vspace{0.5cm}
  \subfigure[Mel-Spectrograms of LibriTTS]{
    \includegraphics[width=0.24\textwidth]{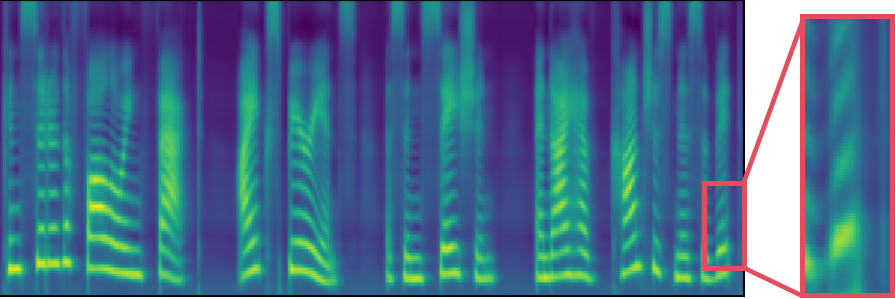}
    \label{subfig31_fl}
    \includegraphics[width=0.24\textwidth]{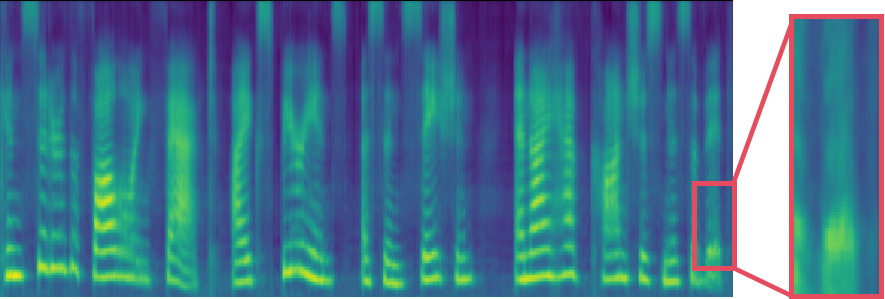}
    \label{subfig32_fl}
    \includegraphics[width=0.24\textwidth]{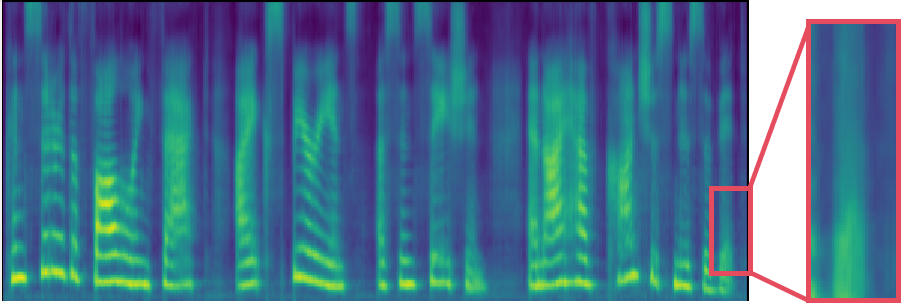}
    \label{subfig33_fl}
    \includegraphics[width=0.24\textwidth]{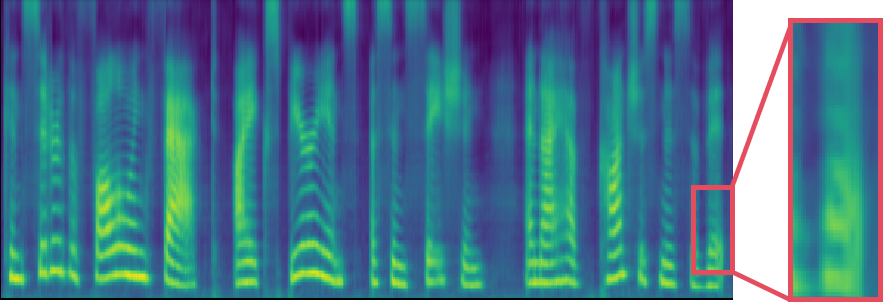}
    \label{subfig34_fl}
  }
  \medskip
  \vspace{0.5cm}
  \subfigure[Mel-Spectrograms of AIshell3]{
    \includegraphics[width=0.24\textwidth]{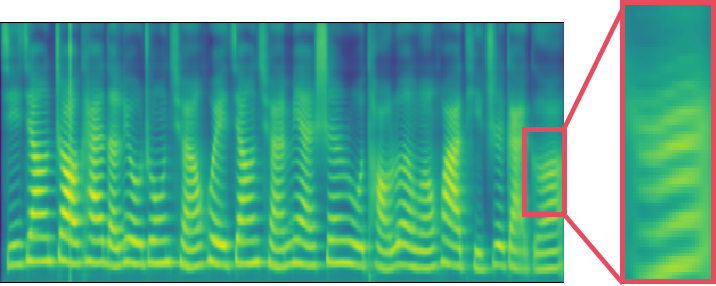}
    \label{subfig41_fl}
    \includegraphics[width=0.24\textwidth]{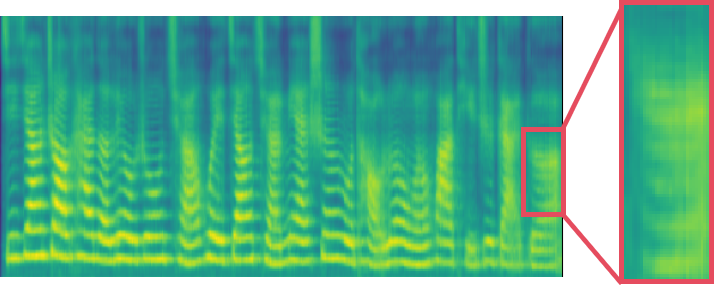}
    \label{subfig42_fl}
    \includegraphics[width=0.24\textwidth]{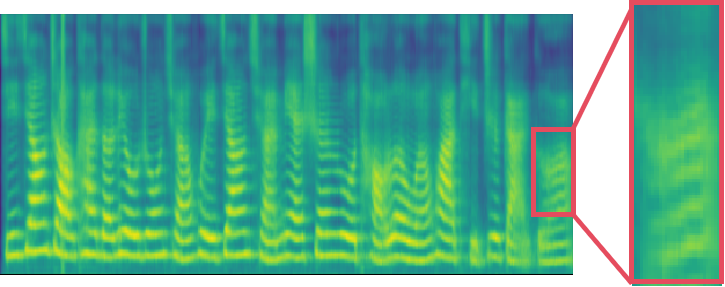}
    \label{subfig43_fl}
    \includegraphics[width=0.24\textwidth]{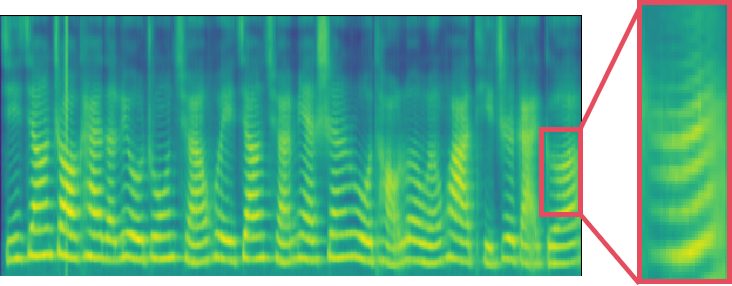}
    \label{subfig44_fl}
  }
  \caption{Mel Spectrograms on LJSpeech, Baker, LibriTTS and Aishell3. Each row from left to right is the Mel spectrograms of the model ANN, SpikeVoice-ATTN, SpikeVoice-SDSA and SpikeVoice-STSA.}
  \label{Mel demo appendix}
\end{figure*}

In Fig.\ref{Mel demo appendix} we present Mel-Spectrograms of LJSpeech, Baker, LibriTTS, and AISHELL3, and we have magnified the tail of the Mel-Spectrogram for a clearer observation.

\end{document}